\documentclass[10pt,twocolumn,letterpaper]{article}

\usepackage[pagenumbers]{cvpr}        %

\usepackage{cuted}    %
\usepackage{currfile} %
\usepackage{caption}  %
\usepackage{multirow}
\usepackage{pifont}
\usepackage{xspace}
\usepackage{amssymb}
\newcommand{\makecell}[1]{\begin{tabular}[c]{@{}c@{}}#1\end{tabular}}
\usepackage{enumitem}
\usepackage{algorithm}
\usepackage{algpseudocode}

\makeatletter
\def\LN@posrec#1#2{\expandafter\gdef\csname LN@x@#1\endcsname{#2}}
\AtEndDocument{\let\LN@posrec\@gobbletwo}
\newcount\LN@seq
\newcommand*\makeLineNumberOuterColumn{%
  \global\advance\LN@seq\@ne
  \pdfsavepos
  \edef\LN@tmp{\write\@auxout{\noexpand\string\noexpand\LN@posrec
    {\the\LN@seq}{\noexpand\the\noexpand\pdflastxpos}}}\LN@tmp
  \expandafter\ifx\csname LN@x@\the\LN@seq\endcsname\relax
    \if@firstcolumn\makeLineNumberLeft\else\makeLineNumberRight\fi
  \else
    \ifdim\csname LN@x@\the\LN@seq\endcsname sp<.5\paperwidth
      \makeLineNumberLeft\else\makeLineNumberRight\fi
  \fi}
\AtBeginDocument{%
  
  }
\makeatother

\definecolor{cvprblue}{rgb}{0.21,0.49,0.74}
\usepackage[breaklinks,colorlinks,allcolors=cvprblue]{hyperref}

\newcommand{\method}{ObjMimic\xspace}
\newcommand{\cpwm}{CP-WM\xspace}
\newcommand{\rgi}{RGI\xspace}
\newcommand{\cmark}{\ding{51}}
\newcommand{\xmark}{\textcolor{black!40}{\ding{55}}}

\hypersetup{pdftitle={Amortizing Object-Trajectory Tracking with a Differentiable Contact World Model},pdfauthor={Zhewen He, Junyi Hu, Jiazhao Liang, Haomian Huang, Shuaihang Yuan, Yi Fang}}

\def\confName{CVPR}
\def\confYear{2027}

\title{Amortizing Object-Trajectory Tracking\\with a Differentiable Contact World Model}

\author{
Zhewen He \quad Junyi Hu \quad Jiazhao Liang \quad Haomian Huang \quad Shuaihang Yuan \quad Yi Fang\thanks{Corresponding author.}\\
New York University
}

\makeatletter
\patchcmd{\@maketitle}{\vspace*{24pt}}{\vspace*{14pt}}{}{\errmessage{title patch 1 failed}}
\patchcmd{\@maketitle}{\vspace*{12pt}}{\vspace*{2pt}}{}{\errmessage{title patch 2 failed}}
\makeatother
\begin{document}
\maketitle
\begin{strip}
  \centering
  \includegraphics[width=\textwidth]{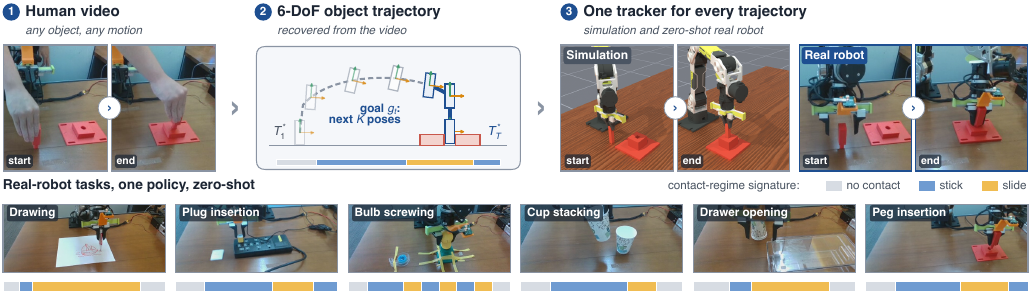}
  \captionsetup{hypcap=false}
  \captionof{figure}{\textbf{One policy for any object trajectory.}
  A 6-DoF object trajectory recovered from a human video is followed by one policy, trained through one differentiable contact world model, in simulation and zero-shot on a real arm. Bottom: six real-robot tasks with the contact-regime signature of each execution.
  }
  \captionsetup{hypcap=true}
  \label{fig:teaser}
\end{strip}

\begin{abstract}
Teaching a robot arm a large library of contact-rich skills typically requires repeated model-free reinforcement learning, whose cost grows with every new task.
We cast manipulation as object-trajectory imitation, where a single policy must make an object follow a target 6-DoF trajectory, and ask whether one differentiable world model can amortize policy learning across tens of thousands of such trajectories.
We introduce \method, whose contact-patch world model operates on partial point clouds and couples contacts through friction-cone-constrained impulses. A learned reliability gate controls analytic policy gradients near contact transitions, without requiring a differentiable simulator.
Trajectories are drawn from a procedural generator with controlled contact regimes and from 6-DoF object motions recovered from human--object interaction videos.
On these trajectories, \method reaches 78.2\% tracking success and matches a fixed target success rate with 12.7 times fewer simulator steps than PPO in the same simulator, and its training cost grows far more slowly with the number of trajectories than that of PPO, rising mainly with the number of distinct contact regimes.
The resulting policy tracks trajectories of unseen objects and transfers to a real arm.
\end{abstract}

\section{Introduction}
\label{sec:intro}

A robot arm that can make any object follow any motion would turn a video of a person handling an object into a skill: recover how the object moves, then reproduce that motion.
Pushing a box, pivoting it onto its side, placing a mug, inserting a peg and opening a drawer differ in their contacts, yet the desired object motion in each can be specified by a 6-DoF trajectory.
Learning such a tracker means learning thousands of contact-rich skills, and the standard recipe, model-free RL with PPO~\citep{schulman2017ppo}, pays for every new skill with new samples~\citep{liu2025dextrack,mandi2025dexmachina}.

Recent work on humanoid tracking offers a related approach.
SuperMimic~\citep{liao2026supermimic} learns one differentiable world model of the shared dynamics of 129.5k human motions from a standard simulator and trains a single tracker by backpropagating through short imagined rollouts, cutting GPU-hours by three orders of magnitude.
The contact in humanoid tracking is, however, between the feet and the ground.
Manipulation is about contact with an external object whose shape the robot only partially sees, that sticks and slides against fingers, supports and constraints, and whose analytic gradients are known to be unreliable near contact transitions~\citep{suh2022diffsim,georgiev2024ahac}.
Whether one world model can amortize policy learning across thousands of such interactions, and what then determines the cost, is open.

We study this question with \method, which casts contact-rich manipulation as object-trajectory imitation (\cref{fig:teaser}).
The policy is conditioned on the next ten reference object poses and receives a dense tracking reward. We quantify library size by the number of reference trajectories.
To ask what drives the cost, we introduce a measurable contact-regime signature: the run-length-encoded sequence of stick, slide and no-contact modes between gripper, object, support and environment.
The signature lets us vary the number of trajectories $N$ and the number of distinct regimes $R$ independently.

Two components make first-order policy learning work in this setting.
The Contact-Patch World Model (\cpwm) operates on partial point clouds: patch tokens fixed in the object frame form candidate contact edges with the gripper and the scene, an edge transformer couples the contacts on each body, and impulses respect the friction cone by construction.
During imagination the geometry moves rigidly with the predicted pose, so no point cloud is re-encoded and each step costs a few hundred edge updates.
Reliability-Gated Imagination (\rgi) learns, from finite-difference probes of the black-box simulator, a geometry-conditioned estimate $\rho$ of whether the model's gradient points the right way.
We turn $\rho$ into a stochastic-truncation objective in which gradients stop after unreliable steps and the critic takes over; full-horizon BPTT is a special case and the adaptive horizon of AHAC~\citep{georgiev2024ahac} a close analogue.
Neither component requires a differentiable simulator.
Both rest on 3D geometry: representing the object by patches of its observed point cloud, rather than by a pose and a class, lets one model decide contact from surface distances and normals, cover unseen shapes, and make its reliability depend on where contact happens.

Trajectories come from a procedural generator with controlled regimes and from 6-DoF object motions recovered from human--object interaction videos.
In ManiSkill3~\citep{tao2024maniskill3}, \method reaches the target success rate with a 12.7-fold reduction in simulator steps relative to PPO in the same simulator.
With a shared world model, its training cost grows far more slowly with the number of trajectories than that of PPO and rises mainly with the number of contact regimes; the policy tracks unseen objects and video-derived trajectories, and transfers to a real arm.

Our contributions are:
\begin{itemize}[leftmargin=*,itemsep=1pt,topsep=2pt]
  \item We formulate contact-rich manipulation as object-trajectory imitation and introduce a measurable contact-regime signature, enabling a controlled study of how policy-learning cost scales with the number of trajectories versus the number of distinct interaction regimes.
  \item We propose the Contact-Patch World Model, a geometry-grounded differentiable dynamics model on partial point clouds that couples multiple contacts through edge attention, keeps impulses friction-cone-feasible by construction and propagates geometry rigidly during imagination.
  \item We introduce Reliability-Gated Imagination, which learns a geometry-conditioned estimate of model-gradient reliability from finite-difference probes of a black-box simulator, and derive a stochastic-truncation objective that unifies full-horizon BPTT and critic bootstrapping.
  \item On procedural and video trajectories, \method reduces simulator steps by a factor of 12.7 relative to PPO, with cost growing mainly with the number of contact regimes. The tracker transfers to unseen objects and a real robot arm.
\end{itemize}

\section{Related Work}
\label{sec:related}

\paragraph{First-order policy learning with differentiable models.}
SHAC~\citep{xu2022shac} showed that analytic gradients through a parallel differentiable simulator train policies with far fewer samples than model-free RL, and AHAC~\citep{georgiev2024ahac} traced its gradient error to stiff contact dynamics and shortened the horizon accordingly.
Both require a differentiable simulator and its contact signals.
Learned models remove that requirement: Dreamer~\citep{hafner2020dreamer,hafner2025dreamerv3} backpropagates through a latent model, SuperMimic~\citep{liao2026supermimic} amortizes humanoid tracking over 129.5k motions with one world model fit from a standard simulator, and Amigo~\etal unroll trajectories in the simulator while taking gradients from a learned model~\citep{amigo2025dmo}, later pairing a diffusion world model with a light latent surrogate~\citep{amigo2026coupled}.
OrbiSim~\citep{li2026orbisim} learns an object-centric differentiable physics engine with a neural renderer, and SGPS~\citep{liu2026sgps} counters convergence of first-order policy gradients to unintended contact patterns with sampling-based MPC refinement.
TD-MPC2~\citep{hansen2024tdmpc2} trains one model on 80 tasks but optimizes actions with MPPI planning and a latent value rather than through model derivatives; PWM~\citep{georgiev2025pwm} backpropagates through such a latent model.
Our focus is on combining explicit contact geometry with learned gradient reliability, and studying how policy-learning cost scales across manipulation trajectories.

\paragraph{Learned contact dynamics and 3D world models.}
ContactNets~\citep{pfrommer2020contactnets} learns discontinuous contact with implicit distances and complementarity constraints.
Graph networks model interacting particles and rigid bodies~\citep{battaglia2016interaction,sanchezgonzalez2020gns,li2019dpinet}, and PIN-WM~\citep{li2025pinwm} embeds physical parameters in a world model for non-prehensile manipulation.
At scale, PointWorld~\citep{huang2026pointworld} predicts full-scene 3D point flows from about 2M trajectories and uses them for model-predictive control.
These models target prediction or planning.
\cpwm is designed for policy gradients: friction-cone feasibility holds by construction, contacts on a body are coupled by attention, and geometry moves rigidly through imagination without re-encoding.

\paragraph{Object-trajectory tracking and learning from video.}
DexTrack~\citep{liu2025dextrack}, DexMachina~\citep{mandi2025dexmachina} and ManipTrans~\citep{li2025maniptrans} train dexterous hands to follow human hand--object references with PPO, using data flywheels, decaying virtual object controllers or residual learning, and SimToolReal~\citep{kedia2026simtoolreal} conditions one object-centric RL policy on goal poses of tools.
RIGVid~\citep{patel2026rigvid} extracts 6-DoF object trajectories from generated videos with FoundationPose~\citep{wen2024foundationpose} and retargets them to the end effector under a rigid-grasp assumption.
We use object-trajectory tracking as a scalable task interface that covers non-prehensile as well as grasping interactions without assuming a rigid grasp, and train the universal tracker with a differentiable world model instead of PPO.

\paragraph{Reliability of gradients through contact.}
Suh~\etal~\citep{suh2022diffsim} analyzed bias and variance of first-order gradients through contact and interpolated between first- and zeroth-order estimators; PIPPS~\citep{parmas2018pipps} combines model-gradient estimators according to their variance, and DDCG~\citep{onoda2026ddcg} switches to zeroth-order estimates where a per-step test detects nonsmooth dynamics.
DiffMJX~\citep{paulus2026diffmjx} repairs hard-contact gradients inside the simulator with adaptive integration and adds contacts from distance for pre-contact gradients.
Sobolev training~\citep{czarnecki2017sobolev} supervises model derivatives, and gradient-aware model learning~\citep{doro2020gradient} weights model errors by their effect on the policy.
These estimators use the true model's derivatives or rollout statistics, not contact geometry.
\rgi learns a geometry-conditioned reliability of a learned model's gradients from finite-difference probes of a black-box simulator and truncates backpropagation per step and per sample.

\section{Method}
\label{sec:method}

\begin{figure*}[t]
  \centering
  \includegraphics[width=\textwidth]{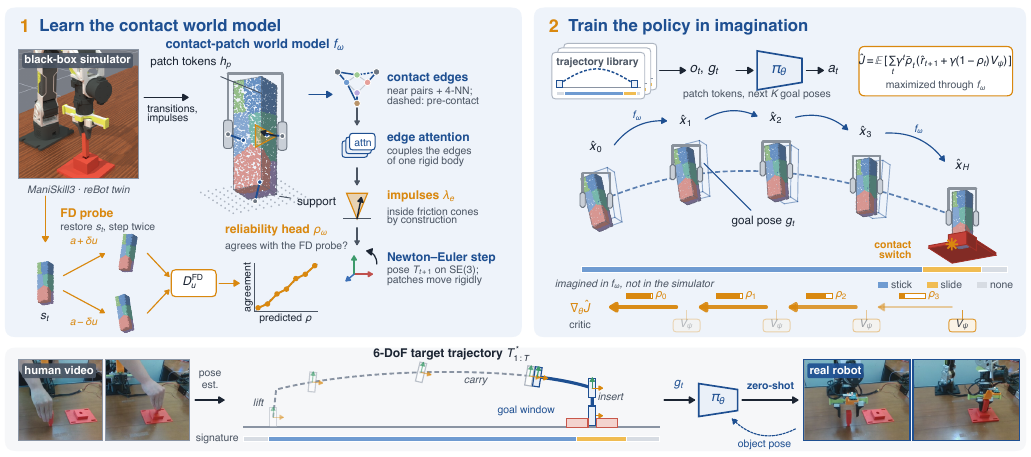}
  \caption{\textbf{Overview of \method.}
  \textbf{(1)}~A contact-patch world model $f_\omega$ is learned from a black-box simulator: the partial point cloud is split into patch tokens $h_p$ fixed in the object frame, contact edges link patches to the fingertips and the support, edge attention couples the edges of each body, impulses $\lambda_e$ stay inside friction cones, and a Newton--Euler step updates the pose on SE(3). A reliability head $\rho_\omega$ learns from finite-difference probes (two perturbed steps from a restored state) whether the model gradient can be trusted.
  \textbf{(2)}~The goal-conditioned policy $\pi_\theta$ tracks trajectories sampled from the library and is trained by maximizing $\hat J$ through rollouts imagined in $f_\omega$ (outlines: goal poses); $\rho_t$ sets how much gradient flows through each step and is lowest at the contact switch, where the critic $V_\psi$ takes over. Strips show the contact-regime signature.
  Bottom: a 6-DoF target trajectory recovered from a human video is tracked zero-shot on the real arm, closed-loop on estimated object poses.
  }
  \label{fig:method}
\end{figure*}

\method trains one goal-conditioned policy to track a target 6-DoF object trajectory.
It alternates three stages, following SuperMimic~\citep{liao2026supermimic}: it collects transitions in a standard, non-differentiable simulator, fits a shared Contact-Patch World Model (\cpwm) to them, and updates the actor and critic by backpropagation through short imagined rollouts of the frozen model.
Two components make this loop work for contact with external objects.
\cpwm (\cref{sec:cpwm}) predicts contact impulses and object motion from partial point clouds, with a structure that keeps its derivatives physically meaningful.
Reliability-Gated Imagination (\rgi, \cref{sec:rgi}) learns where those derivatives can be trusted and hands the remaining steps to the critic.
\Cref{fig:method} gives an overview.

\subsection{Object-Trajectory Imitation}
\label{sec:problem}

The state consists of the arm joint positions $q$, the object pose $T\in\mathrm{SE}(3)$ and the object twist $\xi$.
The policy observes proprioception, the partial object point cloud $P_t$ and the gripper opening, and receives as goal the next $K=10$ reference poses expressed relative to the current pose:
\begin{equation}
\begin{aligned}
  o_t&=\big(q_t,\dot q_t,P_t,\mathrm{grip}_t\big),\qquad
  g_t=\big\{T_t^{-1}T^{*}_{t+k}\big\}_{k=1}^{K},\\
  a_t&=\pi_\theta(o_t,g_t).
\end{aligned}
\label{eq:obs}
\end{equation}
The action is an end-effector delta pose and a gripper command, executed by an operational-space controller~\citep{khatib1987osc} (differential IK on our real arm) and shared by all methods.
The tracking reward multiplies position and rotation terms and penalizes action jumps,
\begin{equation}
  r_t=e^{-\|p_t-p^*_t\|^2/\sigma_p^2}\,e^{-\|\log(R^*_tR_t^\top)\|^2/\sigma_R^2}-c_a\|a_t-a_{t-1}\|^2,
  \label{eq:reward}
\end{equation}
with $\sigma_p=3$\,cm and $\sigma_R=0.4$\,rad, where $\log$ is the SO(3) logarithm.
An episode terminates when the position error exceeds 8\,cm or the rotation error exceeds $60^\circ$.

The reward is dense at every step, and pushing, pivoting, pick-and-place, insertion and drawer opening are all object trajectories, so a skill library becomes a number of trajectories $N$ that can grow to 10k--100k.

\subsection{Contact-Regime Signature}
\label{sec:signature}

Trajectory count alone does not say how diverse a library is.
We therefore measure the interaction regimes it contains.
For a successful execution $\tau$, the simulator's pairwise contact log gives, at every step, a mode for three contact pairs: gripper--object ($\mathrm{go}$), object--support ($\mathrm{os}$) and object--environment constraint ($\mathrm{oe}$, \eg a hole wall or a drawer rail).
Each mode is none, stick or slide, with stick and slide separated by a tangential contact speed of 5\,mm/s.
The signature is the run-length encoding of this sequence,
\begin{equation}
  \sigma(\tau)=\mathrm{RLE}_{L_{\min}}\Big(\big\{(m^{\mathrm{go}}_t,m^{\mathrm{os}}_t,m^{\mathrm{oe}}_t)\big\}_{t=0}^{T}\Big),
  \label{eq:signature}
\end{equation}
where runs shorter than $L_{\min}=5$ steps are merged into their neighbours to remove flickering contacts.
The number of regimes $R(\mathcal L)$ of a library $\mathcal L$ is the number of distinct signatures.
For example, a push is $\mathrm{go}$: none$\to$stick while $\mathrm{os}$ slides; a pivot slides at $\mathrm{go}$ about a sticking $\mathrm{os}$ contact; an insertion sticks at $\mathrm{go}$ while $\mathrm{oe}$ goes from none to slide.

A signature depends on how a trajectory is executed, not only on the reference.
In the procedural library (\cref{sec:setup}) the signature is therefore a generation-time control variable, and the execution log only confirms it.
For video-derived trajectories we compute signatures from successful rollouts after training and use them for analysis only.
This makes $R$ measurable and controllable, so the cost of policy learning can be studied as a function of $N$ and $R$ separately.

\subsection{Contact-Patch World Model}
\label{sec:cpwm}

\paragraph{Tokens.}
We accumulate the partial point cloud of the object in the object frame and sample $M=64$ patch centres by farthest-point sampling~\citep{qi2017pointnetpp}.
A PointNet~\citep{qi2017pointnet} encodes the $k=32$ nearest points of each centre into a token $h_p$.
The gripper fingertips contribute $G=8$ tokens from their known meshes, and the support surface and fixed scene parts contribute tokens from their known geometry.

\paragraph{Candidate contact edges.}
An edge $e=(p,q)$ connects two tokens of different bodies whose centres are closer than $r_c=2$\,cm.
Each object patch also keeps edges to its four nearest tokens on other bodies, which carry gradient before contact.
The edge feature is
\begin{equation}
  z_e=\big(d_e,\,n_e,\,v^{\mathrm{rel}}_e,\,h_p,\,h_q,\,\xi^{\mathrm{phys}}\big),
  \label{eq:edge}
\end{equation}
with signed distance $d_e$, normal $n_e$ and relative velocity $v^{\mathrm{rel}}_e$.
The simulation context $\xi^{\mathrm{phys}}$ (friction, mass) is given to the world model and the critic, never to the actor.

\paragraph{Coupling multiple contacts.}
A four-layer edge transformer of width 128 lets each edge attend to the other edges on the same rigid body.
This approximates the coupling of contact forces that a linear complementarity solver resolves, for example the force balance of a two-finger grasp, which edges predicted independently cannot represent.

\paragraph{Friction-cone impulses by construction.}
From the transformer output, a head predicts a normal logit $w_e$, a gate $s_e\in[0,1]$, a friction estimate $\hat\mu_e$ and a tangential direction $u_e$, and forms the impulse
\begin{equation}
\begin{aligned}
  \lambda_n&=\mathrm{softplus}(w_e)\,s_e,\qquad
  \lambda_\tau=\hat\mu_e\,\lambda_n\tanh(\|u_e\|)\,\tfrac{u_e}{\|u_e\|},\\
  \lambda_e&=\lambda_n n_e+B_e\lambda_\tau,
\end{aligned}
\label{eq:impulse}
\end{equation}
where $B_e$ spans the tangent plane.
The construction guarantees $\lambda_n\ge0$ and $\|\lambda_\tau\|\le\hat\mu_e\lambda_n$ for every input, so all predicted contact impulses remain within their friction cones without a penalty.
The simulator reports pairwise contact impulses, which supervise $\lambda_e$ and the predicted contact mode $\hat m_e$ directly.

\paragraph{Object dynamics.}
The object twist and pose follow a Newton--Euler step on SE(3),
\begin{equation}
\begin{aligned}
  \xi_{t+1}&=\xi_t+\hat M_o^{-1}\Big(\Delta t\,w_g+\textstyle\sum_{e\in E_t}J_e^\top\lambda_e\Big)+\delta_\omega,\\
  T_{t+1}&=T_t\exp\!\big(\Delta t\,\xi_{t+1}^{\wedge}\big),
\end{aligned}
\label{eq:dynamics}
\end{equation}
where the inverse inertia $\hat M_o^{-1}$ is predicted from the patch tokens and $\xi^{\mathrm{phys}}$, $w_g$ is gravity, $J_e$ maps the impulse at the edge to the object wrench, and $\delta_\omega$ is a small learned residual.
A residual MLP predicts the controller-dominated arm dynamics from the commanded action and summed fingertip reaction impulses.
An articulated object is a set of rigid parts whose twists are projected onto the joint degrees of freedom.

\paragraph{Rigid geometry propagation.}
Patch tokens are fixed in the object frame.
In an imagined rollout each step transforms the patch centres by $T_{t+1}$ and recomputes the candidate edges, without re-encoding any point cloud.
A step therefore updates only the candidate edges and involves no rendering, and the geometry in imagination stays consistent with the predicted pose.

\paragraph{Training.}
\cpwm is trained on $8$-step autoregressive rollouts with state, reward and termination terms, plus contact and derivative supervision:
\begin{equation}
\begin{aligned}
  \mathcal L_{\mathrm{WM}}={}&\textstyle\sum_{j=1}^{8}\big[\|\hat x_j-x_j\|^2+(\hat r_j-r_j)^2+\mathrm{BCE}(\hat d_j,d_j)\big]\\
  &+\lambda_m\mathrm{CE}(\hat m_e,m_e)+\lambda_\lambda\|\hat\lambda_e-\lambda^{\mathrm{sim}}_e\|^2\\
  &+\lambda_J\mathcal L_{\mathrm{Sob}}+\lambda_\rho\mathcal L_\rho,
\end{aligned}
\label{eq:wmloss}
\end{equation}
where $d_j$ is the termination flag, $\mathcal L_{\mathrm{Sob}}$ matches directional derivatives within a contact mode (Sobolev training~\citep{czarnecki2017sobolev}) and $\mathcal L_\rho$ trains the reliability head (\cref{sec:rgi}).

\paragraph{Pre-contact gradients.}
Following the contacts-from-distance idea of DiffMJX~\citep{paulus2026diffmjx}, an edge with $d_e>0$ has zero impulse in the forward pass and a straight-through~\citep{bengio2013straight} soft impulse $\kappa_s\,\mathrm{softplus}(-d_e/\epsilon)$ in the backward pass.
The explicit edge representation makes this straightforward to implement and provides gradients that move the gripper toward the object before contact.

\subsection{Reliability-Gated Imagination}
\label{sec:rgi}

Model derivatives through contact can point the wrong way even when the predicted states are accurate~\citep{suh2022diffsim,georgiev2024ahac}.
\rgi learns, from the black-box simulator, where the derivatives of \cpwm agree with the true local response, and uses this estimate inside the policy objective.

\paragraph{Finite-difference probes.}
During data collection, each environment step triggers a probe with probability $p=2\%$.
The probe restores the simulator state, applies $a\pm\delta u$ for a random unit direction $u$ with $\delta=0.05$ in the normalized action space, steps once from each, and records the central difference $D^{\mathrm{FD}}_u$ of the target quantities (object pose increment and reward).
Each probe costs two extra simulator steps, an overhead of 4\% of collection, and the simulator stays a black box.

\paragraph{Reliability head.}
A probe is labelled reliable when the model's directional derivative points the same way as the finite difference and both perturbations stay in the same contact mode:
\begin{equation}
\begin{aligned}
  c&=\mathbb 1\Big[\cos\big(\tfrac{\partial f_\omega}{\partial a}u,\,D^{\mathrm{FD}}_u\big)>\kappa\ \wedge\ m^{+}=m^{-}\Big],\\
  \mathcal L_\rho&=\mathrm{BCE}\big(\rho_\omega(z_{E_t},a_t),\,c\big),
\end{aligned}
\label{eq:rho}
\end{equation}
with $\kappa=0.5$; $c$ is recomputed with the current $f_\omega$ whenever $\rho_\omega$ is trained.
The model derivative is a Jacobian--vector product, which costs about one forward pass.
The head $\rho_\omega$ reads the current edge features $z_{E_t}$, so its estimate is conditioned on contact geometry; it can learn, for instance, that gradients are unreliable while the object is pressed against a side wall.

\paragraph{Stochastic-truncation objective.}
We read $\rho_t$ (with stopped gradient) as the probability of continuing to trust the model after step $t$, and take the expected return under the resulting random truncation of the imagined rollout of length $H$:
\begin{equation}
\begin{aligned}
  \hat J(\theta)&=\mathbb E\Big[\textstyle\sum_{t=0}^{H-1}\gamma^t\hat c_t\,\bar\rho_t\\
  &\quad\cdot\big(\hat r_{t+1}+\gamma(1-\rho_t)V_\psi(\hat x_{t+1},g_{t+1})\big)\\
  &\quad+\gamma^H\hat c_H\,\bar\rho_H\,V_\psi(\hat x_H,g_H)\Big],
\end{aligned}
\label{eq:rgi}
\end{equation}
where $\bar\rho_t=\prod_{k<t}\rho_k$ and $\hat c_t$ is a soft survival weight from the predicted termination.
The policy gradient is the analytic gradient of $\hat J$ through \cpwm.
Three special cases, derived in Supp.~A.1, place the objective among existing estimators.
With $\rho\equiv1$ it is $H$-step backpropagation through time with a terminal critic, the SuperMimic update.
With $\rho\equiv0$ it is a one-step model gradient plus critic, a value-gradient update in the style of SVG(1)~\citep{heess2015svg}.
With $\rho_t\in\{0,1\}$ set by a stiffness threshold it mirrors the adaptive horizon of AHAC~\citep{georgiev2024ahac}.
\rgi instead uses a continuous, per-sample $\rho$ learned from a black-box simulator, so no differentiable simulator is required.
The critic is trained with TD($\lambda$) targets~\citep{sutton1988td} under the same truncation, which keeps the two objectives consistent.

\subsection{Training Loop}
\label{sec:loop}

Each iteration collects transitions with probes in 4096 parallel environments at 20\,Hz, fits \cpwm on a replay buffer of recent transitions, and then freezes it to update the actor and critic with \rgi over imagined rollouts of horizon $H=32$ (1.6\,s) that start from replayed states.
Rollouts are conditioned on trajectories sampled from the whole library, so every iteration covers all regimes.
The actor receives proprioception, the patch tokens of the current partial point cloud and the goal $g_t$; the critic additionally receives the privileged state and $\xi^{\mathrm{phys}}$.
In simulation the point cloud comes from rendered RGB-D with ground-truth segmentation, accumulated in the object frame over the episode so that faces seen earlier remain available after they become occluded.
On the real robot, SAM~2~\citep{ravi2024sam2} and FoundationPose~\citep{wen2024foundationpose} give mask and pose from a static camera, refined near contact from a wrist camera (Supp.~E).
Supp.~A.2 gives the training algorithm, network details and hyperparameters.

\section{Experiments}
\label{sec:experiments}

\begin{table*}[t]
  \centering
  \caption{\textbf{Object-trajectory tracking.}
  Success rate (mean$\pm$std over seeds) on five splits and GPU-hours to $S^*$; all methods share action space, observations, reward and GPU, and PPO-Sim, PPO-WM and \method change one factor at a time.
  \method reaches $S^*$ at a fraction of the cost and leads on the four held-out splits. Best in bold, second underlined.}
  \label{tab:main}
  \small
  \setlength{\tabcolsep}{5pt}
  \begin{tabular}{@{}lllcccccc@{}}
    \toprule
    Method & Dynamics & Update & \makecell{Seen\\SR (\%)\,$\uparrow$} & \makecell{Novel-Traj\\SR (\%)\,$\uparrow$} & \makecell{Novel-Obj\\SR (\%)\,$\uparrow$} & \makecell{Novel-Regime\\SR (\%)\,$\uparrow$} & \makecell{Video\\SR (\%)\,$\uparrow$} & \makecell{GPU-h\\to $S^*$\,$\downarrow$} \\
    \midrule
    PPO-Sim~\citep{schulman2017ppo} & simulator & PPO & 74.0{\scriptsize\,\ensuremath{\pm}\,1.6} & 62.5{\scriptsize\,\ensuremath{\pm}\,1.6} & 47.8{\scriptsize\,\ensuremath{\pm}\,1.7} & 41.9{\scriptsize\,\ensuremath{\pm}\,1.9} & 44.2{\scriptsize\,\ensuremath{\pm}\,1.2} & 2.1k \\
    TD-MPC2~\citep{hansen2024tdmpc2} & latent & MPPI + Q & 70.1{\scriptsize\,\ensuremath{\pm}\,2.1} & 59.3{\scriptsize\,\ensuremath{\pm}\,1.1} & 41.0{\scriptsize\,\ensuremath{\pm}\,0.3} & 37.2{\scriptsize\,\ensuremath{\pm}\,1.7} & 40.3{\scriptsize\,\ensuremath{\pm}\,1.2} & 3.0k \\
    PPO-WM & \cpwm & PPO & 68.6{\scriptsize\,\ensuremath{\pm}\,0.2} & 62.1{\scriptsize\,\ensuremath{\pm}\,1.2} & \underline{48.3}{\scriptsize\,\ensuremath{\pm}\,0.2} & \underline{42.6}{\scriptsize\,\ensuremath{\pm}\,3.1} & \underline{47.9}{\scriptsize\,\ensuremath{\pm}\,3.2} & 830 \\
    MLP-WM + BPTT~\citep{liao2026supermimic} & MLP & BPTT & \textbf{79.4}{\scriptsize\,\ensuremath{\pm}\,1.2} & \underline{63.6}{\scriptsize\,\ensuremath{\pm}\,1.3} & 45.2{\scriptsize\,\ensuremath{\pm}\,0.9} & 37.8{\scriptsize\,\ensuremath{\pm}\,0.2} & 42.7{\scriptsize\,\ensuremath{\pm}\,0.5} & \underline{449} \\
    \method (Ours) & \cpwm & \rgi & \underline{78.2}{\scriptsize\,\ensuremath{\pm}\,0.4} & \textbf{74.7}{\scriptsize\,\ensuremath{\pm}\,1.0} & \textbf{61.8}{\scriptsize\,\ensuremath{\pm}\,0.9} & \textbf{57.8}{\scriptsize\,\ensuremath{\pm}\,1.6} & \textbf{60.6}{\scriptsize\,\ensuremath{\pm}\,0.6} & \textbf{207} \\
    \bottomrule
  \end{tabular}
\end{table*}

\begin{figure*}[t]
  \centering
  \includegraphics[width=\textwidth]{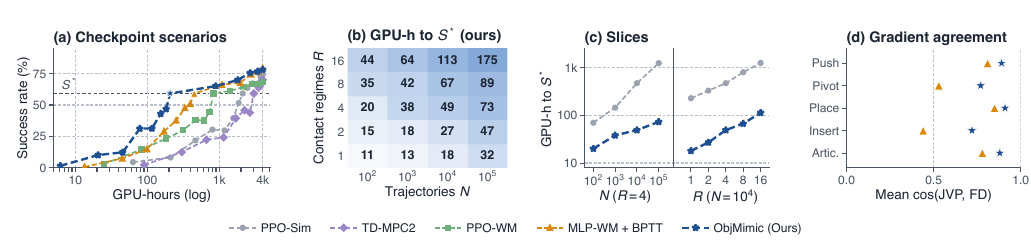}
  \caption{\textbf{Cost, amortization and gradient reliability.}
  (a)~Checkpoints anchored to \cref{tab:main}; dashed segments are guides.
  (b)~GPU-hours of \method to $S^*$ over trajectories $N$ and regimes $R$.
  (c)~\method and PPO-Sim on the slices $R=4$ and $N=10^4$.
  (d)~Mean gradient agreement of each method's world model per family.}
  \label{fig:analysis}
\end{figure*}

\begin{figure*}[t]
  \centering
  \includegraphics[width=\textwidth]{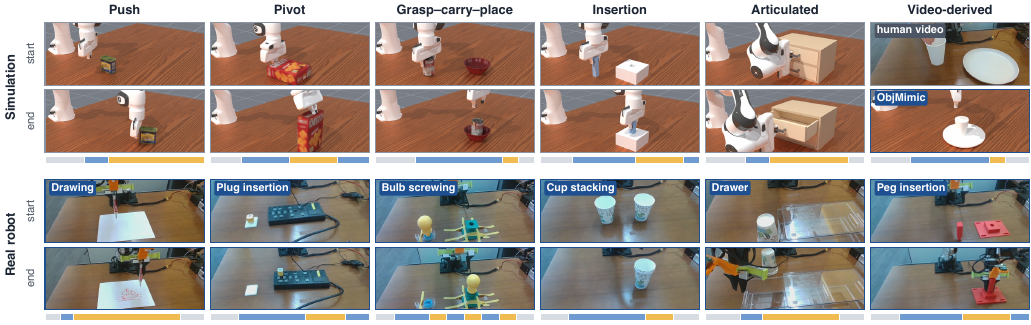}
  \caption{\textbf{Qualitative tracking in simulation and on the real robot.}
  One policy, start and end frames per regime family in simulation (ManiSkill3) and for a trajectory from a human video, and for the six real-robot tasks (bottom); strips show the contact-regime signature. }
  \label{fig:qual}
\end{figure*}

\subsection{Experimental Setup}
\label{sec:setup}

\paragraph{Platform and objects.}
All simulation benchmarks use ManiSkill3~\citep{tao2024maniskill3} with GPU PhysX, a Franka Panda with a parallel gripper, operational-space control at 20\,Hz, and the same GPU type for every method.
Rigid objects come from YCB~\citep{calli2015ycb} and a tabletop subset of Objaverse~\citep{deitke2023objaverse} between 3 and 25\,cm in size; articulated drawers and cabinet doors come from PartNet-Mobility~\citep{xiang2020sapien}.
In total we use 595 objects, holding out 20\% of the instances and 9 whole categories for testing.

\paragraph{Trajectory library.}
The procedural generator produces trajectories in five regime families (push, pivot, grasp--carry--place, insertion, articulated opening), parameterized by displacement, rotation, speed, support surface and friction.
Each trajectory is validated by a privileged oracle (a per-family PPO specialist or an attached-object motion plan), which also fixes its signature (\cref{sec:signature}); the library scales to 100k trajectories.
Video trajectories are 6-DoF object motions from DexYCB~\citep{chao2021dexycb}, HO3D~\citep{hampali2020honnotate}, OakInk2~\citep{zhan2024oakink2}, HOT3D~\citep{banerjee2025hot3d} (rigid) and ARCTIC~\citep{fan2023arctic} (articulated), translated and scaled into the tabletop workspace and filtered for reachability and oracle feasibility, leaving 5.8k trajectories.

\paragraph{Splits.}
\emph{Seen} uses training trajectories.
\emph{Novel-Traj} uses new trajectories on seen objects.
\emph{Novel-Obj} uses held-out instances and categories.
\emph{Novel-Regime} composes known regimes into new signatures, \eg pushing an object against a wall and then pivoting it.
\emph{Video} contains trajectories from held-out video sequences, none of which is used for training; trajectories from the remaining sequences join the training library.

\paragraph{Metrics.}
A rollout succeeds if it completes without early termination and its mean position and rotation errors are below 3\,cm and $15^\circ$, respectively.
Cost is the total GPU-hours to reach a target success rate $S^*$ on the full library, counting collection, probes, world-model fitting and actor--critic updates; $S^*$ is 80\% of the final success rate of PPO-Sim on this library and is kept as the same absolute success rate in every cell of the scaling study.
Gradient quality is measured on steps with contact in two ways: per step, as the cosine between the model's directional derivative and the simulator's finite difference, cos(JVP, FD); and per update, as the cosine between the policy gradient used for training and a finite-difference estimate of the true policy gradient in the simulator, cos($\nabla J$).
We report the mean and standard deviation over 3 seeds.

\paragraph{Implementation.}
\cpwm has about 2.7M parameters and the actor 1.4M.
Every method trains on one node with 8 GPUs of the same type, and GPU-hours count all GPUs, including those used for simulation.
The main comparison trains on $N=10^4$ procedural trajectories sampled to cover all 37 signatures, more than the $R\le16$ of the scaling grid, plus the video training trajectories.

\paragraph{Baselines.}
The main comparison changes one factor at a time along two axes, the source of dynamics and the policy update.
\emph{PPO-Sim} trains PPO~\citep{schulman2017ppo} in the simulator, the standard recipe for multi-trajectory trackers~\citep{liu2025dextrack,mandi2025dexmachina}.
\emph{PPO-WM} trains PPO inside \cpwm, isolating the benefit of fast imagined rollouts.
\emph{MLP-WM + BPTT} replaces \cpwm with the MLP world model of SuperMimic~\citep{liao2026supermimic} and uses ungated BPTT, the direct transfer of SuperMimic to manipulation and, like PWM~\citep{georgiev2025pwm}, first-order learning without a gradient check.
\emph{TD-MPC2}~\citep{hansen2024tdmpc2} is a multi-task model-based baseline that plans with MPPI~\citep{williams2017mppi} using a latent model and a learned Q-function.
All methods share action space, observations, reward, termination, actor size and GPU.
Curriculum PPO in the style of DexMachina, DreamerV3~\citep{hafner2025dreamerv3} and AHAC~\citep{georgiev2024ahac} on a differentiable subset are reported in Supp.~C.5.

\subsection{Main Results}
\label{sec:main}

\Cref{tab:main} and \cref{fig:analysis}(a) compare the five methods.
\method reaches $S^*$ in \textbf{207} GPU-hours, a 10.2-fold reduction relative to PPO-Sim, and leads on the four held-out splits, while MLP-WM + BPTT is slightly better on Seen trajectories.
The two single-factor steps both contribute.
Moving PPO into \cpwm lowers cost through cheaper rollouts, and replacing PPO with \rgi lowers it further, because analytic gradients need far fewer samples than zeroth-order estimates.
MLP-WM + BPTT, the direct transfer of SuperMimic, is competitive on Seen trajectories dominated by sticking grasps but falls behind on Novel-Regime and Novel-Obj, where contact transitions and unseen shapes matter.
TD-MPC2 tracks seen trajectories well but plans at every step, which keeps its cost high, and its latent state generalizes less well to unseen geometry.
\Cref{fig:qual} shows rollouts on unseen objects: the same policy pushes, pivots, places, inserts and opens, and the contact-regime signature of each execution matches its family.

\subsection{Scaling with Trajectories and Regimes}
\label{sec:scaling}

\Cref{fig:analysis}(b) varies the library over $N\in\{10^2,\dots,10^5\}$ and $R\in\{1,\dots,16\}$ with one absolute $S^*$, and \cref{fig:analysis}(c) adds PPO-Sim on two slices (Supp.~C.2).
At $R=4$, a thousandfold larger library raises the cost by 3.7$\times$ for \method and 18$\times$ for PPO-Sim.
At $N=10^4$, it rises from $R=1$ to $16$ by 6.3$\times$ and 5.5$\times$.
The shared world model removes most of the dependence on $N$, while the number of contact regimes drives the cost of both methods.

\begin{table}[t]
  \centering
  \caption{\textbf{Ablation study.}
  Success (\%), policy-gradient agreement cos($\nabla J$) and GPU-hours to $S^*$; ``A$\rightarrow$B'' replaces A by B.
  Ground-truth state improves Seen success and cost; learned gating matters most on new regime combinations.}
  \label{tab:ablation}
  \footnotesize
  \setlength{\tabcolsep}{2.5pt}
  \begin{tabular}{@{}l@{\hspace{1pt}}ccccc@{}}
    \toprule
    Variant & \makecell{Seen\\SR\,$\uparrow$} & \makecell{Nov.-Obj\\SR\,$\uparrow$} & \makecell{Nov.-Reg.\\SR\,$\uparrow$} & \makecell{cos\\($\nabla J$)\,$\uparrow$} & \makecell{GPU-h\\to $S^*$\,$\downarrow$} \\
    \midrule
    \textbf{\method (full)} & \underline{78.2} & \textbf{61.8} & \textbf{57.8} & \textbf{0.59} & \underline{207} \\
    \midrule
    \multicolumn{6}{@{}l}{\emph{Structure}} \\
    \quad w/o edge coupling & 75.9 & 57.1 & 48.9 & 0.45 & 307 \\
    \quad cone $\rightarrow$ penalty & 76.2 & 57.7 & 52.2 & 0.49 & 304 \\
    \quad rigid $\rightarrow$ re-encode & 77.1 & \underline{60.8} & \underline{57.0} & \underline{0.57} & 282 \\
    \quad w/o pre-contact grad. & 75.2 & 58.5 & 54.8 & 0.53 & 268 \\
    \multicolumn{6}{@{}l}{\emph{Input}} \\
    \quad points $\rightarrow$ GT state & \textbf{79.2} & 42.3 & 52.3 & 0.56 & \textbf{189} \\
    \multicolumn{6}{@{}l}{\emph{Reliability}} \\
    \quad $\rho\rightarrow$ ensemble & 75.7 & 59.2 & 50.6 & 0.48 & 306 \\
    \quad $\rho\rightarrow$ hard trunc. & 75.2 & 58.1 & 52.4 & 0.53 & 286 \\
    \quad $\rho\rightarrow1$ (no gating) & 74.4 & 56.9 & 44.7 & 0.35 & 400 \\
    \bottomrule
  \end{tabular}
\end{table}

\subsection{Ablation Study and Gradient Reliability}
\label{sec:ablation}

\Cref{tab:ablation} removes or replaces one component at a time.

\paragraph{World-model structure.}
Predicting each edge's impulse independently loses the force balance of multi-finger contacts and lowers the agreement of the policy gradient to 0.45.
A friction-cone penalty instead of the construction lets imagined rollouts drift outside the cone, which the policy exploits.
Re-encoding the point cloud at every imagined step reaches similar accuracy at a higher cost, and removing the pre-contact gradient slows learning on trajectories that start without contact.

\paragraph{Input.}
Replacing the partial point cloud with the ground-truth object state slightly improves Seen success and lowers cost, but reduces Novel-Obj success to 42.3\%, because a pose alone carries no shape for contact reasoning on new objects.

\paragraph{Reliability.}
Ungated BPTT ($\rho\equiv1$) degrades most on Novel-Regime, where insertion and pivoting cross many contact transitions.
Ensemble disagreement~\citep{chua2018pets} measures value uncertainty, not derivative agreement, which \cref{fig:analysis}(d) compares across contact families.
Hard truncation at contact also discards useful gradients during sticking.
Expected calibration error~\citep{guo2017calibration} averages 0.064 across families (\cref{tab:supp-gradient}).

\subsection{World-Model Analysis}
\label{sec:wm}

\begin{table}[t]
  \centering
  \caption{\textbf{World-model accuracy and throughput.}
  Position error (cm) after 8 and 32 imagined steps, contact-mode F1, impulse error (\%) and steps per second at batch 16k (not ranked).
  Rigid propagation trades a small accuracy loss against re-encoding for higher throughput.}
  \label{tab:wm}
  \footnotesize
  \setlength{\tabcolsep}{2.5pt}
  \begin{tabular}{@{}lccccc@{}}
    \toprule
    Model & \makecell{Pos.\ err.\\8 steps\,$\downarrow$} & \makecell{Pos.\ err.\\32 steps\,$\downarrow$} & \makecell{Mode\\F1\,$\uparrow$} & \makecell{Impulse\\err.\,$\downarrow$} & \makecell{Steps/s\\($\times10^6$)\,$\uparrow$} \\
    \midrule
    \emph{PhysX (simulator)} & -- & -- & -- & -- & 0.40 \\
    \midrule
    MLP-WM~\citep{liao2026supermimic} & 1.97 & 8.15 & -- & -- & 5.85 \\
    \cpwm, re-encode & \textbf{1.01} & \textbf{3.15} & \textbf{0.892} & \underline{13.6} & 0.69 \\
    \cpwm & \underline{1.06} & \underline{3.32} & \underline{0.886} & \textbf{13.3} & 1.69 \\
    \bottomrule
  \end{tabular}
\end{table}

\Cref{tab:wm} evaluates the world models directly.
\cpwm keeps the 32-step position error at 3.32\,cm against 8.15\,cm for the MLP model, and it predicts the contact mode of each edge with an F1 of 0.886.
Re-encoding slightly improves position error and contact-mode F1 but lowers throughput to 0.69M steps/s.
\cpwm steps 4.2$\times$ faster than PhysX state stepping, but rendered simulator steps dominate training, so the gap in \cref{tab:main} comes mainly from needing 1.4B instead of 17.8B of them (Supp.~C.3).

\subsection{Real-Robot Evaluation}
\label{sec:real}

\begin{table}[t]
  \centering
  \caption{\textbf{Real-robot tracking.}
  Successes out of 10 trials per task on a reBot arm.}
  \label{tab:real}
  \small
  \setlength{\tabcolsep}{4pt}
  \begin{tabular}{@{}lccc@{}}
    \toprule
    Task & PPO-Sim\,$\uparrow$ & \makecell{MLP-WM\\+ BPTT\,$\uparrow$} & \method\,$\uparrow$ \\
    \midrule
    Drawing & 5/10 & 6/10 & \textbf{9/10} \\
    Plug & 2/10 & 1/10 & \textbf{6/10} \\
    Bulb & 1/10 & 1/10 & \textbf{4/10} \\
    Cups & 5/10 & 6/10 & \textbf{8/10} \\
    Drawer & 6/10 & 7/10 & \textbf{9/10} \\
    Peg & 3/10 & 2/10 & \textbf{7/10} \\
    \midrule
    Total & 22/60 & 23/60 & \textbf{43/60} \\
    \bottomrule
  \end{tabular}
\end{table}

All methods are retrained on a simulated twin of our real arm, a 6-DoF reBot B601 with a static RGB-D and a wrist camera, and run zero-shot.
The six tasks of \cref{tab:real} are long-horizon and contact-rich; each follows one object trajectory recovered from a human video, with 10 trials per method (Supp.~E).
\method succeeds in 43 of 60 trials, against 22 for PPO-Sim (\cref{tab:real}).
Bulb screwing is hardest: its two-turn helix forces regrasps, where most failures of \method occur.

\section{Conclusion}
\label{sec:conclusion}

\method tracks 6-DoF object trajectories with one policy trained through a differentiable contact world model, with model gradients weighted by a reliability estimate learned from finite-difference probes.
Its cost grows far more slowly than PPO's with the number of trajectories and mainly with the number of contact regimes.

\paragraph{Limitations.}
Success on unseen objects, regime combinations and video trajectories stays near 60\%, well below that on seen trajectories.
Training needs a simulator that reports contact impulses and restores states, so the reliability head cannot be refined on hardware.
Deployment relies on object pose estimates, and deformable and multi-object contact are out of scope (Supp.~F).

{
  \small\setlength{\bibsep}{-0.6pt}
  \bibliographystyle{ieeenat_fullname}
  \bibliography{references}
}

\clearpage
\appendix
\renewcommand{\thetable}{S\arabic{table}}
\renewcommand{\thefigure}{S\arabic{figure}}
\renewcommand{\theequation}{S\arabic{equation}}
\setcounter{table}{0}
\setcounter{figure}{0}
\setcounter{equation}{0}
\makeatletter
\setlength{\@fptop}{0pt}\setlength{\@fpsep}{14pt}\setlength{\@fpbot}{0pt plus 1fil}
\setlength{\@dblfptop}{0pt}\setlength{\@dblfpsep}{14pt}\setlength{\@dblfpbot}{0pt plus 1fil}
\makeatother
\section*{Supplementary Material}

\section{Method Details}
\label{sec:supp-method}

\subsection{Stochastic-Truncation Objective}
\label{sec:supp-derivation}

\paragraph{Statement.}
We state the objective of Eq.~(9) as an expectation over a random truncation and show that its gradient estimator interpolates between existing ones.

\noindent\textbf{Proposition 1.}
\emph{Let $\rho_t\in[0,1]$ be fixed (stop-gradient) and let $\mathcal T$ be the stopping time defined below. Then $\hat J(\theta)=\mathbb E_{\mathcal T}[G_{\mathcal T}]$, and $\nabla_\theta\hat J(\theta)=\mathbb E_{\mathcal T}[\nabla_\theta G_{\mathcal T}]$, where $G_{\mathcal T}$ is the imagined return truncated at $\mathcal T$ and bootstrapped with $V_\psi$.}

\paragraph{Derivation.}
Consider an imagined rollout $\hat x_0,\dots,\hat x_H$ under the policy, with rewards $\hat r_{t+1}$ and survival weights $\hat c_t$.
Let $\mathcal T\in\{0,\dots,H-1\}\cup\{H\}$ be a random stopping time: after receiving $\hat r_{t+1}$, the rollout stops trusting the model with probability $1-\rho_t$, independently of the other steps, and $\mathcal T=H$ if it never stops.
The truncated return bootstraps with the critic at the stopping point,
\begin{equation}
\begin{aligned}
  G_{\mathcal T}={}&\textstyle\sum_{t=0}^{\min(\mathcal T,H-1)}\gamma^t\hat c_t\hat r_{t+1}\\
  &+\gamma^{\mathcal T+1}\hat c_{\mathcal T+1}V_\psi(\hat x_{\mathcal T+1})\,\mathbb 1[\mathcal T<H]\\
  &+\gamma^H\hat c_HV_\psi(\hat x_H)\,\mathbb 1[\mathcal T=H].
\end{aligned}
\end{equation}
Step $t$ is reached with probability $\Pr[\mathcal T\ge t]=\prod_{k<t}\rho_k=\bar\rho_t$, and the rollout stops exactly at $t$ with probability $\bar\rho_t(1-\rho_t)$.
Taking the expectation over $\mathcal T$ term by term,
\begin{equation}
\begin{aligned}
  \mathbb E_{\mathcal T}[G_{\mathcal T}]={}&\textstyle\sum_{t=0}^{H-1}\gamma^t\bar\rho_t\big(\hat c_t\hat r_{t+1}+\gamma(1-\rho_t)\hat c_{t+1}V_\psi(\hat x_{t+1})\big)\\
  &+\gamma^H\bar\rho_H\hat c_HV_\psi(\hat x_H),
\end{aligned}
\end{equation}
which is the objective of Eq.~(9) in the main paper (there $\hat c_{t+1}$ inside the bracket is absorbed into $\hat c_t$, which differs only by the survival of one step).
Since $\rho$ carries a stopped gradient, $\nabla_\theta\hat J$ is the expectation of the analytic gradient of the truncated return: the estimator backpropagates through the model only along steps the rollout still trusts, and the critic's value gradient enters where trust ends.
We compute the expectation in closed form rather than sampling $\mathcal T$, which removes the sampling variance of the stopping time.
The second claim of the proposition follows because $\mathcal T$ does not depend on $\theta$ once $\rho$ is detached, so the expectation over $\mathcal T$, a finite weighted sum, commutes with the gradient. \hfill$\square$

\paragraph{Bias and variance.}
Let $g^{\mathrm{model}}_t$ denote the contribution of step $t$ to the analytic gradient and $g^{\mathrm{true}}_t$ the corresponding contribution under the true dynamics.
Ungated BPTT sums $\sum_t\gamma^t g^{\mathrm{model}}_t$, so a single step with a wrong-signed Jacobian, typical at contact switches~\citep{suh2022diffsim}, corrupts the gradient of every earlier step through the chain rule.
In \rgi the contribution of steps after $t$ is scaled by $\bar\rho_{t+1}$; when $\rho_t$ is low where $g^{\mathrm{model}}_t$ and $g^{\mathrm{true}}_t$ disagree, the wrong Jacobian of step $t$ still enters once, through $\hat r_{t+1}$ and $V_\psi(\hat x_{t+1})$, but it no longer multiplies the gradients of all later steps; those are replaced by the critic's value gradient, whose bias depends on the critic's accuracy rather than on the model's derivatives beyond the contact switch.
Conversely, where $\rho\approx1$ the estimator keeps the low variance of a full-horizon analytic gradient.
The probe labels make this trade-off data-driven: $\rho$ is trained to predict sign agreement, which is the quantity that decides whether a step helps or hurts the policy update, rather than the magnitude of a prediction error.

\paragraph{Relation to $\lambda$-returns.}
With a constant $\rho_t=\rho$, the weights $\bar\rho_t(1-\rho)=\rho^t(1-\rho)$ are those of a TD($\lambda$) return with $\lambda=\rho$ over imagined steps, so \rgi can be read as a value-gradient analogue of TD($\lambda$) whose trace parameter is predicted per step from contact geometry.

\paragraph{Special cases.}
(i)~If $\rho_t=1$ for all $t$, then $\bar\rho_t=1$ and $1-\rho_t=0$, and the objective becomes $\sum_{t<H}\gamma^t\hat c_t\hat r_{t+1}+\gamma^H\hat c_HV_\psi(\hat x_H)$: $H$-step backpropagation through time with a terminal critic, the update of SuperMimic~\citep{liao2026supermimic}.
(ii)~If $\rho_t=0$ for all $t$, then $\bar\rho_0=1$ and $\bar\rho_t=0$ for $t\ge1$, and the objective becomes $\hat c_0\hat r_1+\gamma\hat c_1V_\psi(\hat x_1)$: a one-step model gradient plus the critic's value gradient, as in SVG(1)~\citep{heess2015svg}.
(iii)~If $\rho_t\in\{0,1\}$ is set to $0$ exactly when the simulator's contact stiffness exceeds a threshold, the rollout is truncated after the first stiff step and bootstrapped there, which mirrors the adaptive horizon of AHAC~\citep{georgiev2024ahac}.
\rgi generalizes (iii) in three ways: $\rho$ is continuous, it is predicted per sample from contact geometry, and it is learned from a black-box simulator instead of read from a differentiable one.

\paragraph{Critic target.}
The critic regresses the TD($\lambda$) return~\citep{sutton1988td} computed on the same imagined rollout with the same truncation weights, so that $V_\psi$ estimates the quantity the actor bootstraps with.

\subsection{Implementation Details}
\label{sec:supp-impl}

\paragraph{Networks.}
The patch encoder is a shared PointNet with hidden widths 64--128 and max pooling over the $k=32$ points of each patch.
The edge transformer has four layers of width 128 with four heads; attention is masked to edges incident to the same rigid body.
The actor and critic are MLPs that read proprioception, the pooled patch tokens, the goal poses and, for the critic, the privileged state and $\xi^{\mathrm{phys}}$; the actor size is identical across all methods of the main comparison.
The reliability head is a two-layer MLP on the mean-pooled transformer outputs of the current edges concatenated with the action.

\begin{table*}[t]
  \centering
  \caption{\textbf{Architecture and hyperparameters of \method.} (a)~All networks use LayerNorm and SiLU activations; the actor architecture is shared by all methods of the main comparison. (b)~Selected hyperparameters are tuned on the Seen split.}
  \label{tab:supp-impl}
  \small
  \setlength{\tabcolsep}{6pt}
  \begin{tabular}{@{}lll@{}}
    \toprule
    Module & Layers & Output \\
    \midrule
    Patch encoder & shared PointNet: MLP 3$\rightarrow$64$\rightarrow$128 on $k=32$ points, max-pool & token $h_p\in\mathbb R^{128}$ per patch \\
    Gripper / support tokens & MLP on fingertip mesh vertices / plane parameters & tokens in $\mathbb R^{128}$ \\
    Edge embedding & MLP on $z_e$ (geometry, two tokens, $\xi^{\mathrm{phys}}$) & $\mathbb R^{128}$ per edge \\
    Edge transformer & 4 layers, 4 heads, width 128, attention masked to edges of the same body & $\mathbb R^{128}$ per edge \\
    Impulse head & MLP 128$\rightarrow$64$\rightarrow$6: $w_e$, $s_e$, $\hat\mu_e$, $u_e\in\mathbb R^2$, mode logits & $\lambda_e$, $\hat m_e$ \\
    Inertia head & MLP on pooled object tokens and $\xi^{\mathrm{phys}}$ & $\hat M_o^{-1}$ (SPD, Cholesky) \\
    Arm model & residual MLP on $(q,\dot q,a)$ and summed fingertip impulses & $q_{t+1},\dot q_{t+1}$ \\
    Reward / termination heads & MLP on pooled state & $\hat r$, $\hat d$ \\
    Reliability head & 2-layer MLP on pooled edge outputs and $a_t$ & $\rho_t\in[0,1]$ \\
    Actor & MLP 3$\times$512 on proprioception, pooled patch tokens, $g_t$ & $a_t$ (tanh-squashed) \\
    Critic & MLP 3$\times$768 on actor inputs, privileged state, $\xi^{\mathrm{phys}}$ & $V_\psi$ \\
    \bottomrule
  \end{tabular}
  \par\smallskip(a) Network architecture
  \par\bigskip
  \begin{minipage}[t]{0.48\textwidth}
    \centering
    \begin{tabular}[t]{@{}ll@{}}
      \toprule
      Hyperparameter & Value \\
      \midrule
      \multicolumn{2}{@{}l}{\emph{Task and reward}} \\
      Goal horizon $K$ & 10 steps \\
      Control frequency & 20\,Hz \\
      $\sigma_p$, $\sigma_R$ & 3\,cm, 0.4\,rad \\
      Action-rate weight $c_a$ & 0.01 \\
      Termination & 8\,cm or $60^\circ$ \\
      Success & 3\,cm and $15^\circ$ mean error \\
      \multicolumn{2}{@{}l}{\emph{Contact-regime signature}} \\
      Stick/slide threshold & 5\,mm/s \\
      Minimum run length $L_{\min}$ & 5 steps \\
      \multicolumn{2}{@{}l}{\emph{\cpwm}} \\
      Patches $M$, points per patch $k$ & 64, 32 \\
      Gripper tokens $G$ & 8 \\
      Contact radius $r_c$ & 2\,cm \\
      Pre-contact neighbours & 4 per patch \\
      Rollout length in $\mathcal L_{\mathrm{WM}}$ & 8 steps \\
      Loss weights $\lambda_m$, $\lambda_\lambda$ & 0.5, 1.0 \\
      Loss weights $\lambda_J$, $\lambda_\rho$ & 0.1, 0.5 \\
      WM learning rate ($\times10^{-4}$) & 3.0 \\
      \bottomrule
    \end{tabular}
  \end{minipage}\hfill
  \begin{minipage}[t]{0.48\textwidth}
    \centering
    \begin{tabular}[t]{@{}ll@{}}
      \toprule
      Hyperparameter & Value \\
      \midrule
      \multicolumn{2}{@{}l}{\emph{\rgi}} \\
      Probe rate $p$ & 2\% \\
      Probe step $\delta$ & 0.05 (normalized) \\
      Agreement threshold $\kappa$ & 0.5 \\
      Imagination horizon $H$ & 32 steps (1.6\,s) \\
      Discount $\gamma$, TD $\lambda$ & 0.99, 0.95 \\
      Actor / critic learning rate ($\times10^{-4}$) & 2.0 \\
      Actor--critic steps per iteration $K_\pi$ & 8 \\
      Imagined rollouts per step & 16384 \\
      \multicolumn{2}{@{}l}{\emph{Training}} \\
      Parallel environments & 4096 \\
      Seeds & 3 \\
      \bottomrule
    \end{tabular}
  \end{minipage}
  \par\smallskip(b) Hyperparameters
\end{table*}

\Cref{tab:supp-impl}(a) lists the modules.
The inverse inertia is predicted through a Cholesky factor so that it is symmetric positive definite.
Edges are rebuilt every step with a uniform-grid neighbour search on the GPU; with $M=64$ patches, eight fingertip tokens and a support plane, a typical state has 248 candidate edges, and at most 646.

\paragraph{Training algorithm.}
\Cref{alg:loop} summarizes the training loop of Sec.~3.5.

\begin{algorithm}[t]
  \caption{Training \method}
  \label{alg:loop}
  \small
  \begin{algorithmic}[1]
    \Require trajectory library $\mathcal L$, simulator, probe rate $p$
    \State initialize $f_\omega,\rho_\omega$ (\cpwm), actor $\pi_\theta$, critic $V_\psi$, buffer $\mathcal B$
    \For{iteration $=1,2,\dots$}
      \State sample $\tau^*\sim\mathcal L$ per environment; roll out $\pi_\theta$
      \State with probability $p$ per step: restore state, step $a\pm\delta u$, store $D^{\mathrm{FD}}_u$
      \State add transitions, impulses $\lambda^{\mathrm{sim}}$ and probes to $\mathcal B$
      \State update $f_\omega,\rho_\omega$ on $\mathcal B$ with $\mathcal L_{\mathrm{WM}}$ \Comment{Eq.~(7)}
      \For{actor--critic step $=1,\dots,K_{\pi}$}
        \State imagine $H$ steps with frozen $f_\omega$ from states in $\mathcal B$
        \State $\theta\leftarrow\theta+\eta\nabla_\theta\hat J(\theta)$ \Comment{Eq.~(9), $\rho$ stop-grad}
        \State update $V_\psi$ on truncated TD($\lambda$) targets
      \EndFor
    \EndFor
  \end{algorithmic}
\end{algorithm}

\paragraph{Probes.}
\Cref{alg:probe} details the finite-difference probe.
Restoring the state uses the simulator's state snapshot, and the two perturbed steps run in spare environments of the same batch so that probes do not stall collection.
We use one random direction per probe; directions are drawn uniformly on the unit sphere of the normalized action space.

\begin{algorithm}[t]
  \caption{Finite-difference probe at state $s_t$}
  \label{alg:probe}
  \small
  \begin{algorithmic}[1]
    \Require state $s_t$, action $a_t$, step $\delta$
    \State $u\sim\mathrm{Unif}(\mathbb S^{d-1})$
    \State $s^{+}\leftarrow\mathrm{step}(\mathrm{restore}(s_t),a_t+\delta u)$, $s^{-}\leftarrow\mathrm{step}(\mathrm{restore}(s_t),a_t-\delta u)$
    \State $D^{\mathrm{FD}}_u\leftarrow\big(y(s^{+})-y(s^{-})\big)/(2\delta)$ \Comment{$y$: pose increment, reward}
    \State $m^{\pm}\leftarrow$ contact modes of $s^{\pm}$ from the contact log
    \State store $(s_t,a_t,u,D^{\mathrm{FD}}_u,\mathbb 1[m^{+}=m^{-}])$ \Comment{$c$ of Eq.~(\ref{eq:rho}) is recomputed with the current $f_\omega$ at training time}
  \end{algorithmic}
\end{algorithm}

\paragraph{Hyperparameters.}
\Cref{tab:supp-impl}(b) lists all settings. Values not fixed by the method are selected on the Seen split and fixed before the generalization splits are evaluated.

\paragraph{Baseline settings.}
PPO-Sim and PPO-WM use the same actor and critic networks and observation encoder as \method, with PPO hyperparameters tuned on the Seen split under the same GPU budget.
TD-MPC2 uses its public implementation with the same observation encoder and action space, and its planning horizon tuned on the Seen split.
MLP-WM + BPTT uses the SuperMimic world model, an MLP on the concatenated state, fed the ground-truth object state plus a global shape embedding of the point cloud, and ungated BPTT with the same horizon.

\section{Benchmark and Evaluation Protocol}
\label{sec:supp-benchmark}

\subsection{Trajectory Library}
\label{sec:supp-library}

\paragraph{Procedural generator.}
For each regime family, the generator samples a start pose on the support and a sequence of waypoints: a planar translation for pushing; a rotation about a support edge for pivoting; lift, transport and place segments for pick-and-place; an approach along the hole axis for insertion; and motion along the joint axis for articulated parts.
Speeds, displacements, rotation angles, support friction and obstacles are sampled within ranges listed with the released code.
A trajectory enters the library only if a privileged oracle executes it within the success threshold, and its signature is taken from that execution.
Composite signatures for the Novel-Regime split chain two families, \eg a push that ends against a wall followed by a pivot.

Obstacles and walls are placed for the Novel-Regime compositions only.

\paragraph{Video trajectories.}
Object poses from the five datasets are resampled to 20\,Hz, translated so that the first pose rests on the table, and uniformly scaled when the workspace requires it.
A trajectory is kept if every pose is reachable by inverse kinematics and the oracle tracks it.
Rigid trajectories use the dataset meshes; ARCTIC parts are matched to PartNet-Mobility drawers and doors of similar size.

\paragraph{Statistics.}
\Cref{tab:supp-library} summarizes the library and the test splits.

\begin{table}[!htbp]
  \centering
  \caption{\textbf{Trajectory library and test splits.} Number of trajectories, objects and distinct contact-regime signatures after feasibility filtering; $^\dagger$DexYCB, HO3D, OakInk2 and HOT3D. Signatures of video trajectories are computed from successful executions. Tab.~1 trains on $10^4$ trajectories sampled from the procedural library; the scaling grid samples its libraries from it too.}
  \label{tab:supp-library}
  \small
  \setlength{\tabcolsep}{4pt}
  \begin{tabular}{@{}lccc@{}}
    \toprule
    Source / split & Trajectories & Objects & Regimes $R$ \\
    \midrule
    \multicolumn{4}{@{}l}{\emph{Training library}} \\
    \quad Procedural & 100k & 459 & 37 \\
    \quad Video, rigid$^\dagger$ & 5.1k & 117 & 21 \\
    \quad Video, ARCTIC & 774 & 10 & 6 \\
    \multicolumn{4}{@{}l}{\emph{Test splits}} \\
    \quad Seen & 2.2k & 459 & 37 \\
    \quad Novel-Traj & 2.0k & 439 & 35 \\
    \quad Novel-Obj & 1.0k & 136 & 30 \\
    \quad Novel-Regime & 718 & 249 & 12 \\
    \quad Video & 1.3k & 70 & 11 \\
    \bottomrule
  \end{tabular}
\end{table}

Video trajectories concentrate on sticking grasps and pushes, while the procedural library covers all families evenly.
The regime count $R$ is stable around the chosen stick/slide threshold and $L_{\min}$: very small $L_{\min}$ counts flickering contacts as separate regimes, and above 3 steps $R$ changes little.
In the procedural library the signature is a control variable chosen when a trajectory is generated, and the oracle execution only confirms it, so the $N\times R$ study varies $R$ by construction; signatures of video trajectories are computed after training from successful executions and only describe that split.

\subsection{Evaluation Protocol}
\label{sec:supp-eval}

\paragraph{Splits.}
Held-out categories are chosen so that every regime family keeps at least one held-out category.

\paragraph{Success and errors.}
Position error is the distance between the object's centre of mass and its reference; rotation error is the angle of $R^*_tR_t^\top$.
Both are averaged over the trajectory, and a rollout succeeds when both averages are below their thresholds (3\,cm, $15^\circ$) and the episode was not terminated.
Continuous errors include failed rollouts up to their termination step, so a policy that loses the object early is penalized through a short completed fraction rather than through a large error at the end.

\paragraph{Cost to target.}
$S^*$ is fixed once per library as 80\% of the final success rate of PPO-Sim on the full library.
For each method we log the success rate on a fixed evaluation subset every 10 GPU-hours and report the first time the curve, smoothed over three evaluations, crosses $S^*$.
A method that does not reach $S^*$ within the budget of 4.0k GPU-hours is reported with that budget.

\paragraph{Seeds and statistics.}
Every training configuration is run with 3 seeds; tables report mean and standard deviation over seeds, and each seed is evaluated on every trajectory of each split once.
The scaling grid uses the same seeds for every cell.

\section{Additional Results}
\label{sec:supp-results}

\subsection{Detailed Tracking Results}
\label{sec:supp-tracking}

\paragraph{Unseen objects.}
\Cref{tab:supp-video} separates unseen instances from unseen categories.
Instances share part geometry with training objects, so the patch tokens transfer directly; categories add new contact features such as handles, rims and thin edges.
The drop from instances to categories is about half as large for \method as for the two baselines, which is consistent with the point-cloud input carrying geometry that a learned object embedding does not.
\begin{table}[!htbp]
  \centering
  \caption{\textbf{Success rate (\%) per regime family} on the Novel-Traj split (Place: grasp--carry--place; Artic.: articulated opening). The gap between first-order and zeroth-order methods is largest for pivoting and insertion, and \method leads in every family.}
  \label{tab:supp-family}
  \small
  \setlength{\tabcolsep}{3.5pt}
  \begin{tabular}{@{}lccccc@{}}
    \toprule
    Method & Push & Pivot & Place & Insert & Artic. \\
    \midrule
    PPO-Sim & 70.7 & 49.5 & 73.5 & 48.9 & \underline{69.9} \\
    TD-MPC2 & 65.0 & 47.9 & 71.2 & 46.7 & 65.6 \\
    PPO-WM & \underline{71.1} & 50.5 & \underline{74.0} & 48.0 & 66.9 \\
    MLP-WM + BPTT & 70.6 & \underline{55.5} & 71.6 & \underline{51.3} & 69.1 \\
    \method & \textbf{79.7} & \textbf{68.4} & \textbf{80.0} & \textbf{68.7} & \textbf{76.7} \\
    \bottomrule
  \end{tabular}
\end{table}

\begin{table*}[t]
  \centering
  \caption{\textbf{Novel-Obj and Video success (\%) by subset.} Unseen categories are harder than unseen instances for every method, and \method keeps the smallest gap between the two. ARCTIC trajectories involve articulated parts and OakInk2 contains long bimanual sequences of which we track one object, which makes both the hardest video sources.}
  \label{tab:supp-video}
  \small
  \setlength{\tabcolsep}{5pt}
  \begin{tabular}{@{}lccccccccc@{}}
    \toprule
    & \multicolumn{2}{c}{Novel-Obj} & \multicolumn{6}{c}{Video} \\
    \cmidrule(lr){2-3}\cmidrule(l){4-9}
    Method & Instances & Categories & DexYCB & HO3D & OakInk2 & HOT3D & ARCTIC & All \\
    \midrule
    \emph{Trajectories} & & & 430 & 200 & 287 & 216 & 146 & 1.3k \\
    \midrule
    PPO-Sim & \underline{54.2} & \underline{38.2} & \underline{49.6} & \underline{46.0} & \underline{39.1} & 44.5 & \underline{35.4} & 44.2 \\
    MLP-WM + BPTT & 51.3 & 36.0 & 48.1 & 45.8 & 35.0 & \underline{45.1} & 34.2 & 42.7 \\
    \method & \textbf{65.2} & \textbf{56.7} & \textbf{65.2} & \textbf{63.7} & \textbf{53.6} & \textbf{63.8} & \textbf{52.0} & 60.6 \\
    \bottomrule
  \end{tabular}
\end{table*}

\Cref{tab:supp-family} breaks Novel-Traj success down by regime family, and \cref{tab:supp-video} breaks the Video split down by source dataset.
Grasp-dominated families (pick-and-place, articulated opening) are tracked well by all methods once the grasp is established, whereas pivoting and insertion, which depend on sliding contacts with the support or the environment, separate the methods most.
The ungated MLP model is no better than PPO-Sim on the grasp-dominated families: once the object moves rigidly with the gripper, a correct gradient adds little over a well-tuned zeroth-order estimate.
On insertion, \method leads the strongest baseline by 17.4 percentage points, the largest margin of any family.

\paragraph{Video sources.}
DexYCB is the easiest source and ARCTIC and OakInk2 are the two hardest for every method.
DexYCB and HO3D contain short grasp-and-move sequences, HOT3D adds longer manipulation in cluttered scenes, and ARCTIC and OakInk2 combine articulated parts or bimanual handovers with fast rotations.
The margin of \method over PPO-Sim is roughly constant across sources, so every source contributes to the gain.

\subsection{Scaling Grid}
\label{sec:supp-grid}

Fig.~3(b) gives the GPU-hours of \method to $S^*$ for every cell of the grid, and Fig.~3(c) the two slices for PPO-Sim.
Every cell is trained with the same seeds and evaluation subset and uses the same absolute $S^*$, the target of the main library, so the cells are directly comparable.
From $N=10^2$ to $10^5$, the cost of \method grows 3.7$\times$ at $R=4$ and at most 4.0$\times$ for any $R$, against 18$\times$ for PPO-Sim at $R=4$; at small $N$ a fixed overhead for world-model warm-up and probe calibration dominates.
Even the smallest library ($N=10^2$, $R=16$) keeps six trajectories per signature.
Per-regime PPO specialists, one per signature, share nothing across regimes and need 14.5$\times$ the GPU-hours of \method at $N=10^4$, $R=16$.
Cost is measured on the training library; generalization beyond it is evaluated by the splits of Tab.~1.

\subsection{Cost Breakdown}
\label{sec:supp-cost}

\method uses 1.4 billion simulator steps and 38.1 billion imagined steps, against 17.8 billion simulator steps for PPO-Sim.
Tab.~3 measures state stepping without rendering; in training, every simulator step also renders the cameras, builds the point cloud and runs the actor.
For PPO-Sim, physics stepping takes 12 GPU-hours, rendering, point-cloud processing and rollout inference 1856, and PPO updates 240.
For \method, simulator collection including probes takes 129 GPU-hours, world-model fitting 35, and imagination with its backward pass and the actor--critic updates 43.
Rendered simulator steps thus dominate both budgets, and the cost gap follows mainly from the number of simulator steps each update rule needs.
Imagined steps skip rendering, so although they outnumber simulator steps, they cost a small share; re-encoding the point cloud at every imagined step slows only this share, which is why it raises the total cost in Tab.~2 by a smaller factor than it lowers throughput in Tab.~3.
Per-family final success is reported in \cref{tab:supp-family}.

\subsection{Additional Ablations}
\label{sec:supp-ablation}

\begin{table}[!htbp]
  \centering
  \caption{\textbf{Additional ablations.} Success rates (\%) and cost for horizon, probe rate, auxiliary losses, a low-dimensional shape descriptor instead of the point cloud, and training without video trajectories.}
  \label{tab:supp-ablation}
  \footnotesize
  \setlength{\tabcolsep}{3pt}
  \begin{tabular}{@{}lccc@{}}
    \toprule
    Variant & \makecell{Seen\\SR}\,$\uparrow$ & \makecell{Novel-Reg.\\SR}\,$\uparrow$ & \makecell{GPU-h\\to $S^*$}\,$\downarrow$ \\
    \midrule
    \textbf{full ($H{=}32$, $p{=}2\%$)} & \textbf{78.2} & \textbf{57.8} & 207 \\
    \midrule
    \multicolumn{4}{@{}l}{\emph{Horizon}} \\
    \quad $H=32\rightarrow8$ & 74.0 & 49.2 & 329 \\
    \quad $H=32\rightarrow16$ & 76.4 & 54.0 & 254 \\
    \quad $H=32\rightarrow64$ & 75.3 & 53.1 & 250 \\
    \multicolumn{4}{@{}l}{\emph{Probe rate}} \\
    \quad $p=2\%\rightarrow0.5\%$ & 76.5 & 53.7 & 238 \\
    \quad $p=2\%\rightarrow5\%$ & 77.8 & \underline{57.5} & 215 \\
    \multicolumn{4}{@{}l}{\emph{Losses}} \\
    \quad w/o Sobolev loss & 77.2 & 55.8 & 238 \\
    \quad w/o impulse supervision & 75.9 & 54.1 & 264 \\
    \multicolumn{4}{@{}l}{\emph{Input and data}} \\
    \quad point cloud $\rightarrow$ shape desc. & 75.8 & 53.3 & 219 \\
    \quad proc.\ + video $\rightarrow$ proc. & \underline{77.9} & 57.4 & 208 \\
    \bottomrule
  \end{tabular}
\end{table}

\Cref{tab:supp-ablation} reports the remaining design choices.
Short horizons lose long-range credit assignment and long horizons accumulate model error, and $H=32$ balances the two.
A lower probe rate leaves $\rho$ poorly calibrated, while a higher one adds collection cost without improving success.
Training without video trajectories leaves the procedural splits unchanged but lowers success on the Video split, which the table does not list, by 5.8 percentage points.

\paragraph{Auxiliary losses and input.}
Dropping impulse supervision hurts more than dropping the Sobolev loss: without simulator impulses, the edge transformer can fit the next pose with impulses that are physically wrong, and their derivatives point the wrong way at contact switches.
Replacing the point cloud by a low-dimensional shape descriptor costs most on Novel-Regime, where the new regimes involve edges and faces that the descriptor cannot localize.

\begin{figure*}[t]
  \centering
  \includegraphics[width=\textwidth]{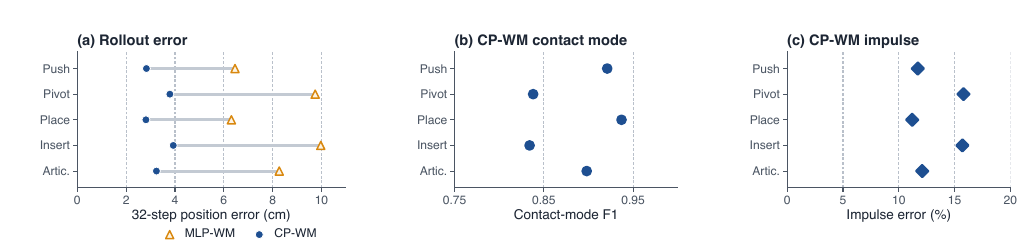}
  \caption{\textbf{World-model errors by contact family.} (a)~32-step position error of MLP-WM and \cpwm. (b)~Contact-mode F1 and (c)~impulse error of \cpwm, on separate axes.}
  \label{fig:supp-wm}
\end{figure*}

\subsection{Additional Baselines}
\label{sec:supp-baselines}

\begin{table}[!htbp]
  \centering
  \caption{\textbf{Additional baselines} on the full library (success in \%). The curriculum is an object-controller curriculum following the decaying virtual object controller of DexMachina~\citep{mandi2025dexmachina}.}
  \label{tab:supp-baselines}
  \small
  \setlength{\tabcolsep}{4pt}
  \begin{tabular}{@{}lccc@{}}
    \toprule
    Method & \makecell{Seen\\SR}\,$\uparrow$ & \makecell{Novel-Obj\\SR}\,$\uparrow$ & \makecell{GPU-h\\to $S^*$}\,$\downarrow$ \\
    \midrule
    PPO-Sim + curriculum & \underline{77.3} & \underline{50.6} & \underline{1.9k} \\
    DreamerV3~\citep{hafner2025dreamerv3} & 67.0 & 40.2 & 2.7k \\
    \method (Ours) & \textbf{78.2} & \textbf{61.8} & \textbf{207} \\
    \bottomrule
  \end{tabular}
\end{table}

\Cref{tab:supp-baselines} adds two baselines.
The object-controller curriculum strengthens PPO-Sim but remains model-free, and DreamerV3 shares the latent-model design of TD-MPC2.
AHAC~\citep{georgiev2024ahac} requires a differentiable simulator and cannot run in ManiSkill3; on a push and pivot subset reimplemented in MJX, it serves as an upper reference for gradient quality rather than for compute, since it differentiates the simulator itself, reaching a cos(JVP, FD) of 0.88 against 0.82 for \method retrained on MJX rollouts of the same subset, within its ManiSkill3 range for push and pivot (\cref{tab:supp-gradient}).
OrbiSim~\citep{li2026orbisim} is not compared because its code is not released.

\paragraph{Gap to the extra baselines.}
The curriculum shortens exploration early in training, but its virtual controller has to be annealed per trajectory, so its cost still grows with the library.
DreamerV3, like TD-MPC2, learns a latent model from the same observations and inherits the same difficulty with contact switches: its imagined returns are accurate in free motion and unreliable through contact.
Its cost lies between those of PPO-Sim and TD-MPC2, since its actor is trained on imagined returns without any check of the model gradient.

\section{Gradient and World-Model Analysis}
\label{sec:supp-analysis}

\subsection{Gradient Analysis}
\label{sec:supp-gradient}

\begin{table}[!htbp]
  \centering
  \caption{\textbf{Gradient agreement per regime family.} Cosine between model directional derivative and finite difference on contact steps, calibration error of $\rho$, and relative change of the true simulator return after one gated gradient step ($\Delta J$). Pivot and insertion, with many contact transitions, gain most from the contact-patch structure.}
  \label{tab:supp-gradient}
  \footnotesize
  \setlength{\tabcolsep}{3pt}
  \begin{tabular}{@{}lcccc@{}}
    \toprule
    Regime family & \makecell{cos\\MLP-WM}\,$\uparrow$ & \makecell{cos\\\cpwm}\,$\uparrow$ & \makecell{ECE\\of $\rho$}\,$\downarrow$ & \makecell{$\Delta J$\\(\%)}\,$\uparrow$ \\
    \midrule
    Push & 0.81 & 0.89 & 0.070 & 1.1 \\
    Pivot & 0.53 & 0.77 & 0.066 & 0.7 \\
    Grasp--carry--pl. & 0.85 & 0.91 & 0.046 & 1.1 \\
    Insertion & 0.44 & 0.72 & 0.052 & 0.9 \\
    Articulated & 0.78 & 0.88 & 0.084 & 1.1 \\
    \bottomrule
  \end{tabular}
\end{table}

\Cref{tab:supp-gradient} lists the gradient means shown in Fig.~3(d), together with calibration error and return change by family.
$\Delta J$ is measured by applying the policy update along $\hat g$ and evaluating the policy in the simulator before and after.
The ungated MLP model agrees with finite differences on push and pick-and-place but falls to 0.44 on insertion, where the object is pressed against the hole wall for most of the trajectory.
\cpwm keeps the cosine at 0.72 or higher in every family.
The calibration error of $\rho$ stays low in all families, so one gate serves every regime.

\subsection{World-Model Analysis}
\label{sec:supp-wm}

\Cref{fig:supp-wm} breaks the accuracy of Tab.~3 down by regime family.
The horizon endpoints and throughput at batch size 16k are reported only in Tab.~3; the throughput protocol includes full edge recomputation at every step.

\paragraph{Errors by family.}
Both contact-mode F1 and impulse error are worse on pivot and insertion than on the grasp-dominated families; the two metrics need not give exactly the same family ranking.
The 32-step position error of the MLP model is more than twice that of \cpwm in every family.
The re-encoding variant in Tab.~3 isolates rigid propagation: it encodes the transformed point cloud at every step, which slightly improves position error and contact-mode F1 but reduces throughput; rigid propagation retains the better impulse error.

\begin{figure*}[t]
  \centering
  \includegraphics[width=\textwidth]{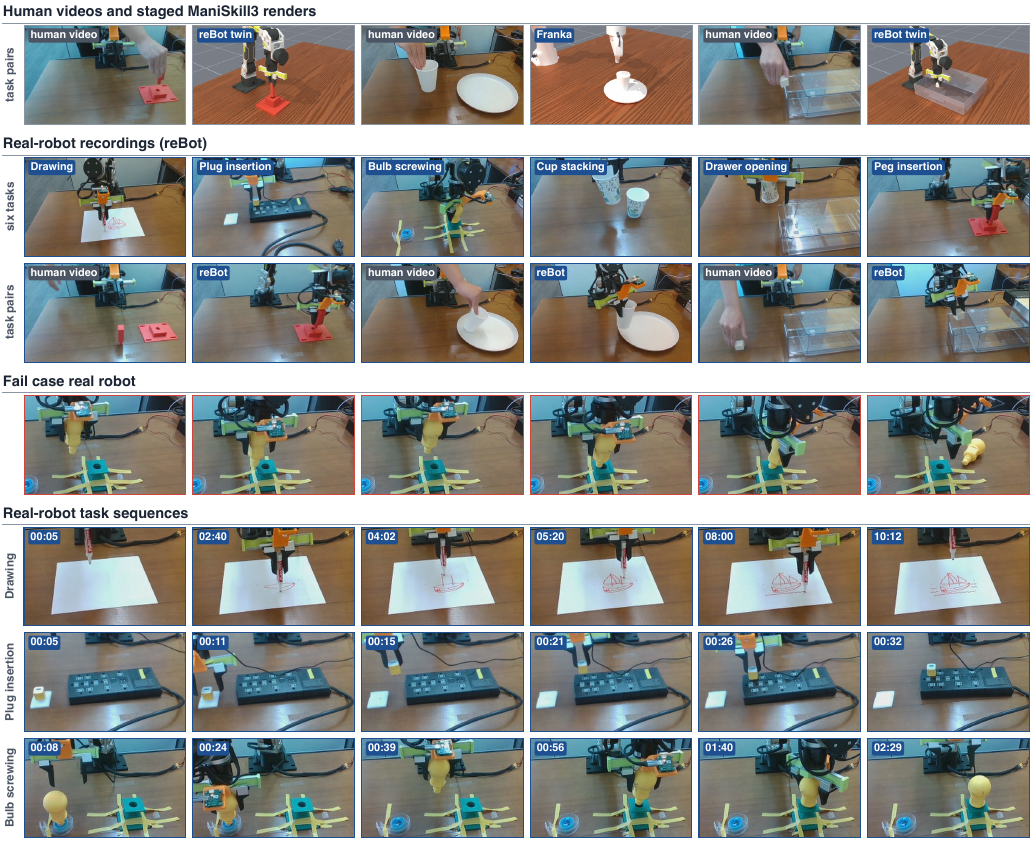}
  \caption{\textbf{Additional qualitative results and failure cases.}
  \emph{Simulation:} Three human videos paired with ManiSkill3 renders of peg insertion and drawer opening (reBot twin) and cup placement (Franka Panda).
  \emph{Real robot:} The six tasks, and the same three human videos paired with reBot recordings of the corresponding tasks.
  The third block shows frames from failed trials on the reBot in which the bulb misses the socket opening, contacts the rim without entering, or falls after release.
  The last block shows six-frame sequences of drawing, plug insertion, and bulb screwing on the reBot.}
  \label{fig:supp-qual}
\end{figure*}

\section{Real-Robot Setup and Results}
\label{sec:supp-real}

All policies are trained in simulation and deployed without real-world fine-tuning. Hardware measurements are used only to identify the simulator's PD gains and latency (below).
This is a requirement of the method rather than a convenience, since the reliability head is supervised by finite-difference probes that restore a simulator state and step it twice, which a physical system cannot do, and the cost argument of the paper rests on thousands of parallel environments.

\paragraph{Hardware and embodiment.}
The real system is a Seeed reBot Arm B601-RS, a 6-DoF arm with a parallel gripper (rated payload 2.5\,kg, repeatability 0.1\,mm), observed by one static Intel RealSense D435 and a wide-angle RGB camera on the wrist.
The simulation benchmark uses a Franka, but the policy, the arm model of \cpwm and the gripper tokens depend on the embodiment, so the Franka policies are not transferred.
Instead, every method is retrained in ManiSkill3 on a twin of the reBot built from its official URDF, with the same networks, hyperparameters and GPU-budget rule as in the benchmark.
The object trajectories need no change, since they are defined on the object alone; we only re-filter the library with the privileged oracle executed by the reBot, which keeps 75\% of the trajectories, and training the reBot policy of \method takes 161 GPU-hours.
The action is still an end-effector delta pose and a gripper command; on the reBot, damped least-squares differential inverse kinematics converts it into joint position targets that the joint actuators track with their internal PD loop, identically in the twin and on the hardware.
The policy runs at 20\,Hz.

\paragraph{Perception.}
SAM~2~\citep{ravi2024sam2} segments the object in the first D435 frame and tracks the mask; FoundationPose~\citep{wen2024foundationpose} estimates the object pose, which places the accumulated point cloud in the object frame and provides $T_t$ for the goal.
Near contact, the fingers and the arm occlude the object from the static camera, which matters most for insertion and for thin objects.
Whenever the object is visible in the wrist image, MegaPose~\citep{labbe2022megapose} refines the pose on that image, initialized with the current estimate and placed in the world frame by forward kinematics and hand--eye calibration, and the refined pose replaces the D435 estimate.
The wrist camera is used only for pose estimation, so the actor receives the same inputs as in simulation.

\paragraph{Tasks.}
The six tasks are chosen to fit the gripper and the payload, and none of the objects appears in training.
They cover sliding on a support (a pen drawing on paper), insertion after grasp and carry (a charger plug, a peg), insertion with twisting and regrasps (a bulb into its socket), placing inside another object (stacking two cups) and articulated sliding (opening a drawer by its handle).
For each task, one human demonstration video is recorded with the static camera, and the 6-DoF object trajectory is recovered with the pipeline of the Video split, restricted to the reBot workspace.
Start poses of the object vary between trials within a few centimetres, and every method starts from the same set of poses.
A trial succeeds when the end state of the task is reached (stroke completed, plug and peg seated, bulb seated in its socket, cup nested, drawer open) and the object stays within the tracking thresholds of simulation.

\paragraph{Sim-to-real measures.}
Three measures close the gap between simulation and the robot.
First, dynamics are randomized per episode (\cref{tab:supp-simreal}(a)).
The sampled friction and mass form the context $\xi^{\mathrm{phys}}$ of Sec.~3.3, which conditions \cpwm and the critic but never the actor, so imagined rollouts cover the whole randomized range and the actor is optimized to track under all of it; the remaining parameters are hidden from every network.
Second, the observation pipeline is reproduced in simulation: the simulated static camera is placed at the calibrated extrinsics of the D435 with its resolution and intrinsics, depth noise and point dropout are applied before patching, and pose noise is added to $T_t$ before it enters the goal, so that the actor never receives information that FoundationPose cannot provide.
Third, two quantities are identified on the hardware rather than randomized blindly: the joint PD gains of the twin are fitted so that simulated joint step responses match those measured on the reBot, and the action latency range covers the measured delay from image capture to joint command, including segmentation and pose estimation.
Training, probes and all simulation evaluations in the paper use these ranges, so the simulation results already include this variation.

\begin{table*}[t]
  \centering
  \caption{\textbf{Sim-to-real randomization and robustness.} (a)~Ranges sampled per episode, uniformly unless a standard deviation is given; multiplicative ranges scale the nominal value. Controller gains are the joint PD gains on the reBot and the operational-space gains on the Franka; the same ranges train every method. (b)~Success in simulation with Gaussian noise on the observed object pose and random dropout of the partial point cloud, the two errors that dominate on the real system; accumulating the point cloud in the object frame makes \method tolerant to dropout.}
  \label{tab:supp-simreal}
  \small
  \setlength{\tabcolsep}{4pt}
  \begin{minipage}[t]{0.475\textwidth}
    \centering
    \begin{tabular}{@{}lll@{}}
        \toprule
        Group & Parameter & Range \\
        \midrule
        Object & friction coefficient & $[0.3,\,1.2]$ \\
         & mass & $\times[0.7,\,1.3]$ \\
         & center-of-mass offset & $\pm1$\,cm per axis \\
         & scale & $\times[0.95,\,1.05]$ \\
        Robot & fingertip friction & $[0.6,\,1.4]$ \\
         & controller gains & $\times[0.8,\,1.2]$ \\
         & action latency & 0--2 steps (0--100\,ms) \\
         & action noise & $\sigma=0.02$ (normalized) \\
        Scene & table friction & $[0.3,\,0.8]$ \\
         & hinge and slide damping & $\times[0.5,\,2.0]$ \\
        Perception & pose noise & $\sigma\le0.5$\,cm\,/\,$2^\circ$ \\
         & point dropout & 0--30\% \\
         & depth noise & $\sigma=2$\,mm \\
         & camera extrinsics & $\pm1$\,cm\,/\,$\pm2^\circ$ \\
        \bottomrule
      \end{tabular}
    \par\smallskip(a) Domain randomization
  \end{minipage}\hfill
  \begin{minipage}[t]{0.505\textwidth}
    \centering
    \begin{tabular}{@{}lccc@{}}
        \toprule
        Perturbation & PPO-Sim\,$\uparrow$ & \makecell{MLP-WM\\+ BPTT}\,$\uparrow$ & \method\,$\uparrow$ \\
        \midrule
        None & 62.5 & \underline{63.6} & \textbf{74.7} \\
        Pose noise 0.5\,cm / $2^\circ$ & 57.7 & \underline{60.4} & \textbf{71.5} \\
        Pose noise 1.0\,cm / $5^\circ$ & \underline{52.0} & 49.6 & \textbf{65.8} \\
        Point dropout 30\% & \underline{57.7} & 56.5 & \textbf{72.4} \\
        Point dropout 60\% & \underline{54.7} & 49.7 & \textbf{71.4} \\
        \bottomrule
      \end{tabular}
    \par\smallskip(b) Novel-Traj success (\%) under perception noise
  \end{minipage}
\end{table*}

\paragraph{Perception noise.}
\Cref{tab:supp-simreal}(b) injects the perception errors of the real system in simulation to separate them from the dynamics gap; the larger pose noise and 60\% dropout lie outside the training ranges of \cref{tab:supp-simreal}(a).
At the larger settings, pose noise is more harmful than point dropout for every method, since it shifts the goal $T_t^{-1}T^*_{t+1:t+K}$ that the policy tracks.
\method loses only 3.3 percentage points at 60\% dropout, because the point cloud accumulated in the object frame keeps the patches it has already seen.
The MLP model, which encodes the whole cloud into one embedding, is the most sensitive to both perturbations.

\paragraph{Sim-to-real gap.}
On the same 60 trials in the reBot twin, \method succeeds in 81.7\% of the trials.
Its success drops by 10.0 percentage points on the hardware, against 13.3 for PPO-Sim.
Failures of \method concentrate on the regrasps of bulb screwing and on plug insertion, where the plug is small and its pose estimate is noisy at insertion depth.

\section{Qualitative Results, Failure Cases and Limitations}
\label{sec:supp-limitations}

\Cref{fig:supp-qual} adds rollouts to Fig.~4 of the main paper, including a video-derived trajectory next to its source frames and a real-robot trial.

The most frequent failures are slips of thin objects under fast rotation, slips during the regrasps of bulb screwing on the real robot, and pivots on geometric features that the 64 sampled patches miss.

\method needs a simulator that reports pairwise contact impulses and can restore states for probes; ManiSkill3 provides both, but not every simulator does.
The contact-regime signature measures diversity only for regimes the simulator resolves, and deformable, granular or multi-object contact is outside the current model.
At deployment the policy depends on object pose estimates from FoundationPose~\citep{wen2024foundationpose}, refined near contact by MegaPose~\citep{labbe2022megapose}; \cref{tab:supp-simreal}(b) shows how tracking degrades with pose noise and point dropout, and tracking failures under heavy occlusion end the episode.
Library construction relies on a privileged oracle, and the success criterion bounds the mean tracking error, which does not by itself verify the end state of insertion or opening; task completion is checked on the real robot.
Success on unseen objects, regime combinations and video trajectories remains around 60\%, well below success on seen trajectories.
The scaling result holds within the tested grid of up to 100k trajectories and 16 regimes.
A single tracker for many trajectories lowers the cost of teaching robots from video, which helps assistive and industrial uses; the human--object datasets we use were collected with consent by their authors, and we release only derived object trajectories where the licenses permit.

\section{Discussion}
\label{sec:supp-discussion}

\subsection{Comparison with Related Work}
\label{sec:supp-related}

\Cref{tab:supp-related} contrasts \method with the closest methods of Sec.~2 on six properties.

\begin{table*}[t]
  \centering
  \caption{\textbf{Comparison with related work.}
  ``Learned diff.\ model'': policy gradients through a learned model; ``No diff.\ simulator'': no differentiable simulator required; ``3D geometry'': point-cloud or asset geometry as model input; ``External contact'': contact with manipulated objects; ``Cost scaling'': a controlled study of how cost grows with the number of tasks; ``Gradient reliability'': an explicit estimate of where gradients through contact can be trusted.
  PointWorld learns a 3D model but uses it for MPC, not policy gradients; TD-MPC2 optimizes with planning and a latent value rather than model derivatives.
  A \xmark{} means the property is not reported in the public material.}
  \label{tab:supp-related}
  \small
  \setlength{\tabcolsep}{6pt}
  \begin{tabular}{@{}lcccccc@{}}
    \toprule
    Method & \makecell{Learned\\diff.\ model} & \makecell{No diff.\\simulator} & \makecell{3D\\geometry} & \makecell{External\\contact} & \makecell{Cost\\scaling} & \makecell{Gradient\\reliability} \\
    \midrule
    SHAC~\citep{xu2022shac} & \xmark & \xmark & \xmark & \xmark & \xmark & \xmark \\
    AHAC~\citep{georgiev2024ahac} & \xmark & \xmark & \xmark & \xmark & \xmark & \cmark \\
    SuperMimic~\citep{liao2026supermimic} & \cmark & \cmark & \xmark & \xmark & \cmark & \xmark \\
    Coupled L--G WM~\citep{amigo2026coupled} & \cmark & \cmark & \xmark & \cmark & \xmark & \xmark \\
    OrbiSim~\citep{li2026orbisim} & \cmark & \cmark & \cmark & \cmark & \xmark & \xmark \\
    PointWorld~\citep{huang2026pointworld} & \xmark & \cmark & \cmark & \cmark & \xmark & \xmark \\
    TD-MPC2~\citep{hansen2024tdmpc2} & \xmark & \cmark & \xmark & \cmark & \xmark & \xmark \\
    PWM~\citep{georgiev2025pwm} & \cmark & \cmark & \xmark & \cmark & \xmark & \xmark \\
    DDCG~\citep{onoda2026ddcg} & \xmark & \xmark & \xmark & \xmark & \xmark & \cmark \\
    DexTrack~\citep{liu2025dextrack} & \xmark & \cmark & \xmark & \cmark & \xmark & \xmark \\
    \method (Ours) & \cmark & \cmark & \cmark & \cmark & \cmark & \cmark \\
    \bottomrule
  \end{tabular}
\end{table*}

\paragraph{SuperMimic on an arm.}
SuperMimic~\citep{liao2026supermimic} tracks human motions whose only contact is between the feet and the ground, with a world model on a flat state vector.
Manipulation adds an external object that the robot sees only partially, contact with several surfaces at once, and frequent switches between sticking and sliding.
The MLP-WM + BPTT row of Tab.~1 is this direct transfer; it trails \method most on Novel-Obj and Novel-Regime, and the contact-patch structure and the reliability gate account for the difference (Tab.~2).

\paragraph{Learned reliability and ensemble disagreement.}
Ensemble disagreement~\citep{chua2018pets} measures uncertainty in predicted values.
A small next-state error alone does not constrain a model's derivative at a contact switch; Fig.~3(d) therefore reports derivative agreement directly.
$\rho$ is trained on exactly this event, sign agreement with a finite difference, and Tab.~2 shows that it gates better than disagreement.

\end{document}